\documentclass[a4paper,11pt]{article}
\usepackage{graphicx}
\usepackage[usenames,svgnames]{xcolor}
\usepackage[setpagesize=false,colorlinks=true,linkcolor=blue,citecolor=blue,breaklinks=true]{hyperref}
\usepackage{amsmath,amssymb,bm,amsthm}

\usepackage{braket}

\usepackage[numbers,sort&compress]{natbib}

\begin{document}

\title{Improved generalization by noise enhancement}
\author{Takashi Mori$^1$ \and Masahito Ueda$^{1,2,3}$}
\date{$^1$\textit{RIKEN Center for Emergent Matter Science (CEMS), Wako, Saitama 351-0198, Japan}\\
$^2$\textit{Department of Physics, Graduate School of Science, The University of Tokyo, Bunkyo-ku, Tokyo 113-0033, Japan}\\
$^3$\textit{Institute for Physics of Intelligence, University of Tokyo, 7-3-1, Hongo, Bunkyo-ku, Tokyo 113-0033, Japan}}
\maketitle

\begin{abstract}
Recent studies have demonstrated that noise in stochastic gradient descent (SGD) is closely related to generalization: A larger SGD noise, if not too large, results in better generalization.
Since the covariance of the SGD noise is proportional to $\eta^2/B$, where $\eta$ is the learning rate and $B$ is the minibatch size of SGD, the SGD noise has so far been controlled by changing $\eta$ and/or $B$.
However, too large $\eta$ results in instability in the training dynamics and a small $B$ prevents scalable parallel computation.
It is thus desirable to develop a method of controlling the SGD noise without changing $\eta$ and $B$.
In this paper, we propose a method that achieves this goal using ``noise enhancement'', which is easily implemented in practice.
We expound the underlying theoretical idea and demonstrate that the noise enhancement actually improves generalization for real datasets.
It turns out that large-batch training with the noise enhancement even shows better generalization compared with small-batch training.
\end{abstract}

\section{Introduction}
It is a big theoretical challenge in deep learning studies to understand why networks trained via stochastic gradient descent (SGD) and its variants generalize so well in the overparameterized regime, in which the number of network parameters greatly exceeds that of the training data samples~\citep{Zhang2017}.
This fundamental problem has been tackled from different points of view~\citep{Dziugaite2017, Nagarajan2017, Neyshabur2017, Neyshabur2019, Arora2018, Perez2019, Jacot2018, Arora2019, dAscoli2020}.
Among them, some recent studies have pointed out the importance of an implicit regularization effect of SGD~\citep{Zhu2019, Wu2019, Smith2020}.
Indeed, it is empirically known that the SGD noise strength is strongly correlated with generalization of the trained network~\citep{Li2017, Jastrzebski2017, Goyal2017, Smith-Le2018, Hoffer2017, Hoffer2019}.
It has also been argued that the SGD noise prefers wide flat minima, which are considered to indicate good generalization~\citep{Keskar2017, Hoffer2017, Wu2018}.
From this viewpoint, not only its strength, but also the structure of the SGD noise is considered to be important since it is theoretically shown that the network can efficiently escape from bad local minima with the help of the SGD noise but not of an isotropic Gaussian noise with the same strength~\citep{Zhu2019, Wu2019}.

The covariance of the SGD noise is proportional to $\eta^2/B$, where $\eta$ and $B$ denote the learning rate and the minibatch size, respectively, and hence, the SGD noise strength can be controlled by changing $\eta$ and/or $B$.
To realize good generalization, we want to increase the SGD noise strength by increasing $\eta$ and/or decreasing $B$.
However, when $\eta$ becomes too large, the training dynamics often becomes unstable and the training fails.
On the other hand, decreasing $B$ prevents an efficient parallelization using multiple GPUs or TPUs\footnote{However, it is not at all trivial whether the large-batch training is really efficient even with an ideal parallelization. See \citet{Golmant2018, Hoffer2019} for scalability of large-batch training.}.
It is therefore desirable to control the SGD noise without changing these hyperparameters.

The main contribution of the present paper is to show that the SGD noise can be controlled without changing $\eta$ and $B$ by a simple yet efficient method that we call \textit{noise enhancement}.
In this method, the gradient of the loss function is evaluated by using two independent minibatches.
We will explain our theoretical idea in Sec.~\ref{sec:NE}.
We will also demonstrate that the noise enhancement improves generalization in Sec.~\ref{sec:experiment}.
In particular, it is empirically shown that the large-batch training using the noise enhancement even outperforms the small-batch training.
This result gives us some insights into the relation between the SGD noise and generalization, which is discussed in Sec.~\ref{sec:discussion}.
Because of its simplicity in implementation, this method would also be useful in practice.

\section{Noise enhancement}
\label{sec:NE}

We shall consider a classification problem.
The training dataset $\mathcal{D}=\{(x^{(\mu)},y^{(\mu)})\}_{\mu=1,2,\dots,N}$ consists of pairs of the input data vector $x^{(\mu)}$ and its label $y^{(\mu)}$.
The set of all the network parameters is simply denoted by $w$.
Then the output of the network for a given input $x$ is denoted by $f(x;w)$.
The loss function is defined as
\begin{equation}
L(w)=\frac{1}{N}\sum_{\mu=1}^N\ell\left(f(x^{(\mu)};w),y^{(\mu)}\right)\equiv\frac{1}{N}\sum_{\mu=1}^N\ell_\mu(w),
\end{equation}
where the function $\ell(\cdot,\cdot)$ specifies the loss (in this paper we employ the cross-entropy loss).

In the SGD, the training data is divided into minibatches of size $B$, and the parameter update is done by using one of them.
Let $\mathcal{B}_t\subset\{1,2,\dots,N\}$ with $|\mathcal{B}_t|=B$ be a random minibatch chosen at the $t$-th step, the network parameter $w_t$ is updated as
\begin{equation}
w_{t+1}=w_t-\eta\nabla_wL_{\mathcal{B}_t}(w_t), \quad L_{\mathcal{B}_t}(w)=\frac{1}{B}\sum_{\mu\in\mathcal{B}_t}\ell_\mu(w_t)
\end{equation}
in vanilla SGD, where $\eta>0$ is the learning rate.
It is also expressed as
\begin{equation}
w_{t+1}=w_t-\eta\nabla_wL(w_t)-\eta\left[\nabla_wL_{\mathcal{B}_t}(w_t)-\nabla_wL(w_t)\right]
\equiv w_t-\eta\nabla_wL(w_t)-\xi_t(w_t).
\label{eq:SGD}
\end{equation}
Here, $\xi_t$ corresponds to the SGD noise since its average over samplings of random minibatches is zero: $\mathbb{E}_{\mathcal{B}_t}[\xi_t]=0$.
Its covariance is also calculated straightforwardly~\citep{Zhu2019}:
\begin{align}
\mathbb{E}_{\mathcal{B}_t}\left[\xi_t\xi_t^\mathrm{T}\right]&=\frac{\eta^2}{B}\frac{N-B}{N-1}\left(\frac{1}{N}\sum_{\mu=1}^N\nabla_w\ell_\mu\nabla_w\ell_\mu^\mathrm{T}-\nabla_wL\nabla_wL^\mathrm{T}\right)
\nonumber \\
&\approx\frac{\eta^2}{B}\left(\frac{1}{N}\sum_{\mu=1}^N\nabla_w\ell_\mu\nabla_w\ell_\mu^\mathrm{T}-\nabla_wL\nabla_wL^\mathrm{T}\right),
\label{eq:noise_var}
\end{align}
where we assume $N\gg B$ in obtaining the last expression.
This expression\footnote{From Eq.~(\ref{eq:noise_var}), some authors~\citep{Krizhevsky2014, Hoffer2017} argue that the SGD noise strength is proportional to $\eta/\sqrt{B}$, while others~\citep{Li2017, Jastrzebski2017, Smith2018} argue that it is rather proportional to $\sqrt{\eta/B}$ on the basis of the stochastic differential equation obtained for an infinitesimal $\eta\to +0$.
Thus the learning-rate dependence of the noise strength is rather complicated.}
shows that the SGD noise strength is controlled by $\eta$ and $B$.

We want to enhance the SGD noise without changing $\eta$ and $B$.
Naively, it is possible just by replacing $\xi_t$ by $\alpha\xi_t$ with a new parameter $\alpha>1$.
Equation~(\ref{eq:SGD}) is then written as
\begin{align}
w_{t+1}&=w_t-\eta\nabla_wL(w_t)-\alpha\xi_t(w_t)
\nonumber \\
&=w_t-\eta\left[\alpha\nabla_wL_{\mathcal{B}_t}(w_t)+(1-\alpha)\nabla_wL(w_t)\right].
\label{eq:naive_NE}
\end{align}
Practically, Eq.~(\ref{eq:naive_NE}) would be useless because the computation of $\nabla_wL(w_t)$, i.e. the gradient of the loss function over the entire training data, is required for each iteration\footnote{If we have computational resources large enough to realize ideal parallelization for full training dataset, this naive noise enhancement would work. However, with limited computational resources, it is not desirable that we have to evaluate $\nabla_wL(w_t)$ for each iteration.}.
Instead, we propose replacing $\nabla_wL(w_t)$ in Eq.~(\ref{eq:naive_NE}) by $\nabla_wL_{\mathcal{B}_t'}(w_t)$, where $\mathcal{B}_t'$ is another minibatch of the same size $B$ that is independent of $\mathcal{B}_t$.
We thus obtain the following update rule of the \textit{noise-enhanced SGD}:
\begin{equation}
w_{t+1}=w_t-\eta\left[\alpha\nabla_wL_{\mathcal{B}_t}(w_t)+(1-\alpha)\nabla_wL_{\mathcal{B}_t'}(w_t)\right].
\label{eq:NE}
\end{equation}
By defining the SGD noise $\xi_t'$ associated with $\mathcal{B}_t'$ as
\begin{equation}
\xi_t'(w_t)=\eta\left[\nabla_wL_{\mathcal{B}_t'}(w_t)-\nabla_wL(w_t)\right],
\end{equation}
Eq.~(\ref{eq:NE}) is rewritten as
\begin{equation}
w_{t+1}=w_t-\eta\nabla_wL(w_t)-\xi_t^\mathrm{NE}(w_t),
\end{equation}
where the noise $\xi_t^\mathrm{NE}$ in the noise-enhanced SGD is given by
\begin{equation}
\xi_t^\mathrm{NE}=\alpha\xi_t+(1-\alpha)\xi_t'.
\end{equation}
Its mean is obviously zero, i.e. $\mathbb{E}_{\mathcal{B}_t,\mathcal{B}_t'}[\xi_t^\mathrm{NE}]=0$, and its covariance is given by
\begin{align}
\mathbb{E}_{\mathcal{B}_t,\mathcal{B}_t'}\left[\xi_t^\mathrm{NE}\left(\xi_t^\mathrm{NE}\right)^\mathrm{T}\right]
&=\alpha^2\mathbb{E}_{\mathcal{B}_t}\left[\xi_t\xi_t^\mathrm{T}\right]+(1-\alpha^2)\mathbb{E}_{\mathcal{B}_t'}\left[\xi_t'(\xi_t')^\mathrm{T}\right]
\nonumber \\
&=\left[\alpha^2+(1-\alpha)^2\right]\mathbb{E}_{\mathcal{B}_t}\left[\xi_t\xi_t^\mathrm{T}\right],
\label{eq:NE_var}
\end{align}
where we have used the fact that two noises $\xi_t$ and $\xi_t'$ are i.i.d. random variables.
In this way, the SGD-noise covariance is enhanced by a factor of $\alpha^2+(1-\alpha)^2>1$ for $\alpha>1$.
Since the size of the new minibatch $\mathcal{B}_t'$ is same as that of the original minibatch $\mathcal{B}_t$, the noise enhancement does not suffer from any serious computational cost.

If we assume $N\gg B$, Eq.~(\ref{eq:NE_var}) is equivalent to Eq.~(\ref{eq:noise_var}) with an effective minibatch size
\begin{equation}
B_\mathrm{eff}=\frac{B}{\alpha^2+(1-\alpha)^2}.
\label{eq:B_eff}
\end{equation}
If the SGD noise were Gaussian, it would mean that the noise-enhanced SGD is equivalent to the normal SGD with the effective minibatch size $B_\mathrm{eff}$.
However, the SGD noise is actually far from Gaussian during training~\citep{Panigrahi2019}, at least for not too large minibatch size. 
The noise enhancement is therefore not equivalent to reducing the minibatch size unless $B_\mathrm{eff}$ is too large.

The procedure of the noise enhancement is summarized as the follows: (i) prepare two independent minibatches $\mathcal{B}_t$ and $\mathcal{B}_t'$, and (ii) replace the minibatch gradient $\nabla_wL_{\mathcal{B}_t}(w_t)$ by $\alpha\nabla_wL_{\mathcal{B}_t}(w_t)+(1-\alpha)\nabla_wL_{\mathcal{B}_t'}(w_t)$.
The numerical implementation is quite simple.
It should be noted that the noise enhancement is also applicable to other variants of SGD like Adam.

\section{Experiment}
\label{sec:experiment}

\begin{table}[t]
\caption{Network configurations.}
\vspace{11pt}
\centering
\begin{tabular}{lllll}
Name & Network type & Dataset & $L^*$ & $L^{**}$ \\
\hline
F1 & Fully connected & Fashion-MNIST & 0.01 & 0.001 \\
C1 & Convolutional & Cifar-10 & 0.01 & 0.001 \\
C2 & Convolutional & Cifar-100 & 0.02 & 0.001\\
\hline
\end{tabular}
\label{t:config}
\end{table}

We shall demonstrate the efficiency of the method of the noise enhancement (NE) for several network configurations with a real dataset as listed in Table~\ref{t:config}.

We describe the details of the network architecture below:
\begin{itemize}
\item\textbf{F1}: A fully-connected feed-forward network with 7 hidden layers, each of which has 500 neurons with the ReLU activation. The output layer consists of 10 neurons with the softmax activation.
\item\textbf{C1}: A modified version of the VGG configuration~\citep{Simonyan2014}.
Following \citet{Keskar2017}, let us denote a stack of $n$ convolutional layers of $a$ filters and a kernel size of $b\times c$ with the stride length of $d$ by $n\times[a,b,c,d]$.
The C1 network uses the configuration: $3\times[64,3,3,1]$, $3\times[128,3,3,1]$, $3\times[256,3,3,1]$, where a MaxPool(2) is applied after each stack.
To all layers, the ghost-batch normalization of size 100 and the ReLU activation are applied.
Finally, an output layer consists of 10 neurons with the softmax activation.
\item\textbf{C2}: It is similar to but larger than C1. The C2 network uses the configuration: $3\times[64,3,3,1]$, $3\times[128,3,3,1]$, $3\times[256,3,3,1]$, $2\times[512,3,3,1]$, where a MaxPool(2) is applied after each stack.
To all layers above, the ghost-batch normalization of size 100 and the ReLU activation are applied.
The last stack above is followed by a 1024-dimensional dense layer with the ReLU activation, and finally, an output layer consists of 10 neurons with the softmax activation.
\end{itemize}

For all experiments, we used the cross-entropy loss and the Adam optimizer with the default hyperparameters.
Neither data augmentation nor weight decay is applied in our experiment.
To aid the convergence, we halves the learning rate when the training loss reaches the value $L^*$.
Training finishes when the training loss becomes smaller than the value $L^{**}$.
Our choices of $L^*$ and $L^{**}$ are also described in Table~\ref{t:config}.
The convergence time is defined as the number of iteration steps until the training finishes.
Training is repeated 10 times starting from different random initializations (the Glorot initizalization is used), and we measure the mean test accuracy and the mean convergence time as well as their standard deviations.

\subsection{Effect of the noise enhancement}

\begin{figure}[t]
\centering
\begin{tabular}{cc}
\includegraphics[width=0.5\linewidth]{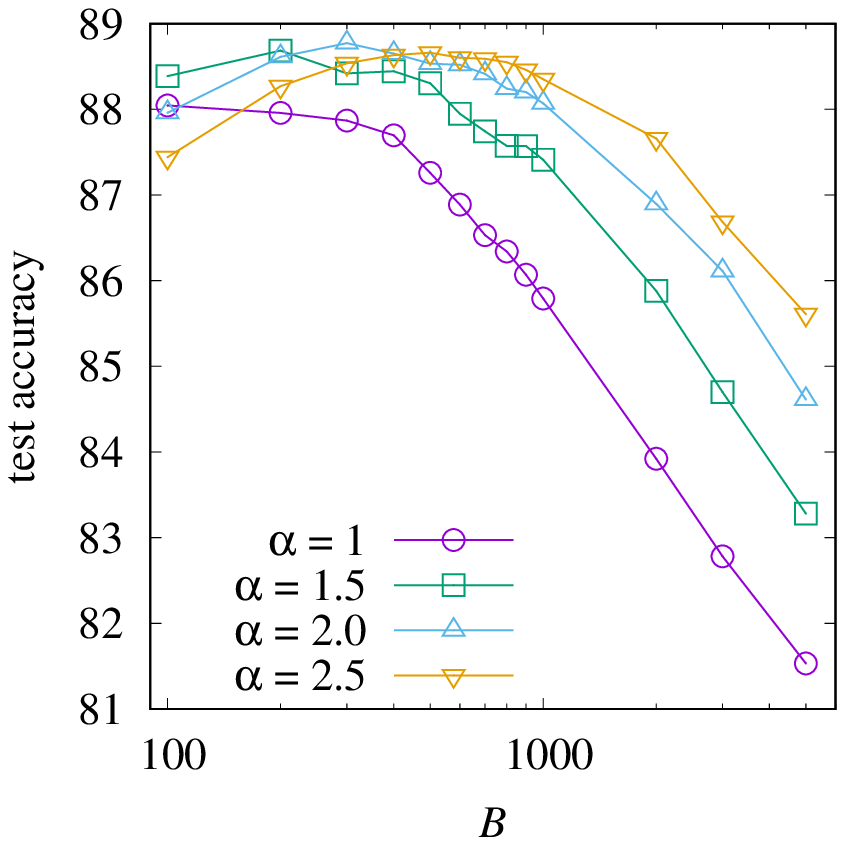}&
\includegraphics[width=0.5\linewidth]{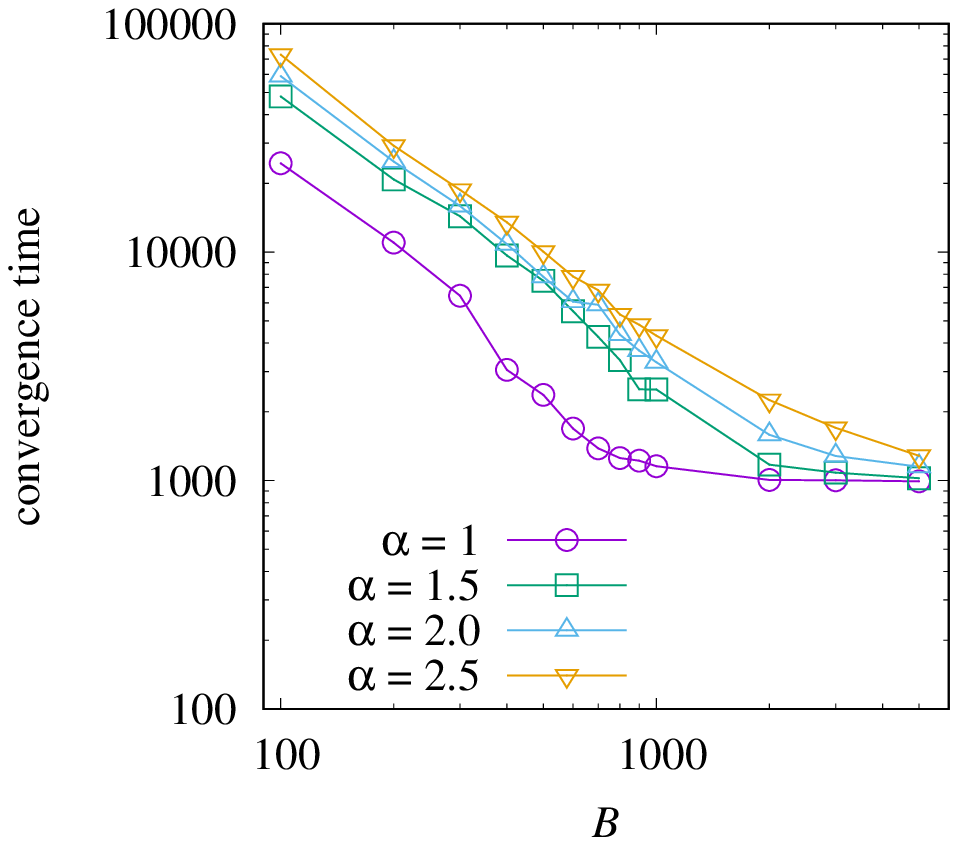}
\end{tabular}
\caption{Minibatch-size dependence of the test accuracy (left) and the convergence time (right) for each fixed value of $\alpha$ in C1.}
\label{fig:batch_dep}
\end{figure}

\begin{table}[t]
\caption{Best test accuracy for each value of $\alpha$.}
\vspace{11pt}
\centering
\begin{tabular}{l|llll}
Name & $\alpha$ & $B_\mathrm{opt}$ & test accuracy (\%) & convergence time \\
\hline\hline
C1
&1 & 100 & $88.05\pm 0.18$ & $24500\pm 2775$ \\
&1.5 & 200 & $88.69\pm 0.11$ & $20825\pm 2113$ \\
&2.0 & 300 & $\bm{88.77\pm 0.30}$ & $15932\pm 1870$ \\
&2.5 & 500 & $88.66\pm 0.22$ & $\bm{10040\pm 1153}$ \\
\hline
F1
&1 & 900 & $90.17\pm 0.14$ & $10934\pm 816$ \\
&1.5 & 2000 & $\bm{90.39\pm 0.17}$ & $\bm{7914\pm 528}$ \\
\hline
C2
&1 & 600 & $61.40\pm 0.54$ & $5292\pm 935$ \\
& 1.5 & 1000 & $\bm{61.75\pm 0.48}$ & $\bm{5175\pm 748}$ \\
\hline
\end{tabular}
\label{t:batch_dep}
\end{table}

First we demonstrate how the noise enhancement affects the generalization and the convergence time for C1 (similar results are obtained for F1 and C2 as we show later).
For each fixed value of $\alpha=1, 1.5, 2.0, 2.5$ ($\alpha=1$ means no NE applied) we calculated the mean test accuracy and the mean convergence time for varying minibatch sizes $B$.
The result is presented in Fig.~\ref{fig:batch_dep}.
We can see that the NE improves generalization for a not too large $\alpha$.
It is also observed that the generalization gap between small-batch training and large-batch training diminishes by increasing $\alpha$.
The NE with large $\alpha$ is therefore efficient for large-batch training.
On the other hand, the convergence time increases with $\alpha$ for a fixed $B$.

For each fixed $\alpha$, there is an optimal minibatch size $B_\mathrm{opt}$, which increases with $\alpha$.
In Table~\ref{t:batch_dep}, we list $B_\mathrm{opt}\in\{100,200,300,400,500,600,700,800,900,1000,2000,3000,5000\}$ as well as the test accuracy and the convergence time at $B=B_\mathrm{opt}$.
We see that the test accuracy at $B_\mathrm{opt}$ is improved by the NE.
Moreover, the NE shortens the convergence time at $B_\mathrm{opt}$ without hurting generalization performance\footnote{The NE for a fixed $B$ increases the convergence time, but $B_\mathrm{opt}$ also increases, which decreases the convergence time.}.
This experimental observation shows practical efficiency of the method of the NE.

\begin{figure}[t]
\centering
\begin{tabular}{cc}
(a) test accuracy for F1 & (b) convergence time for F1 \\
\includegraphics[width=0.5\linewidth]{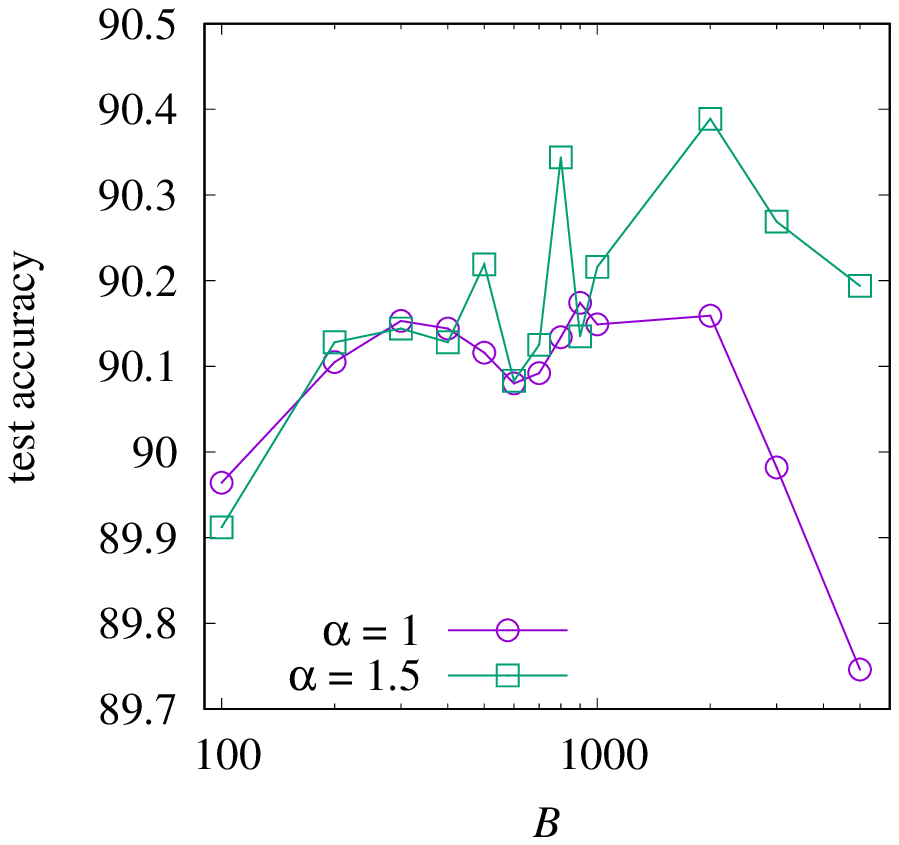}&
\includegraphics[width=0.5\linewidth]{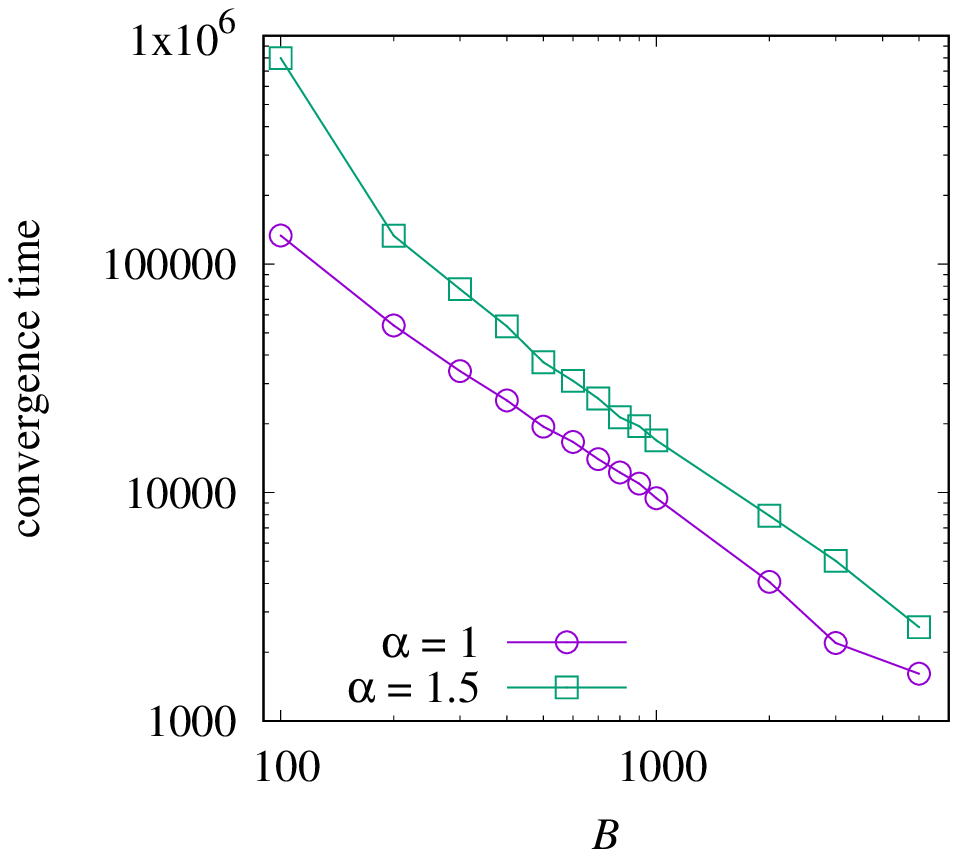}\\
(c) test accuracy for C2 & (d) convergence time for C2 \\
\includegraphics[width=0.5\linewidth]{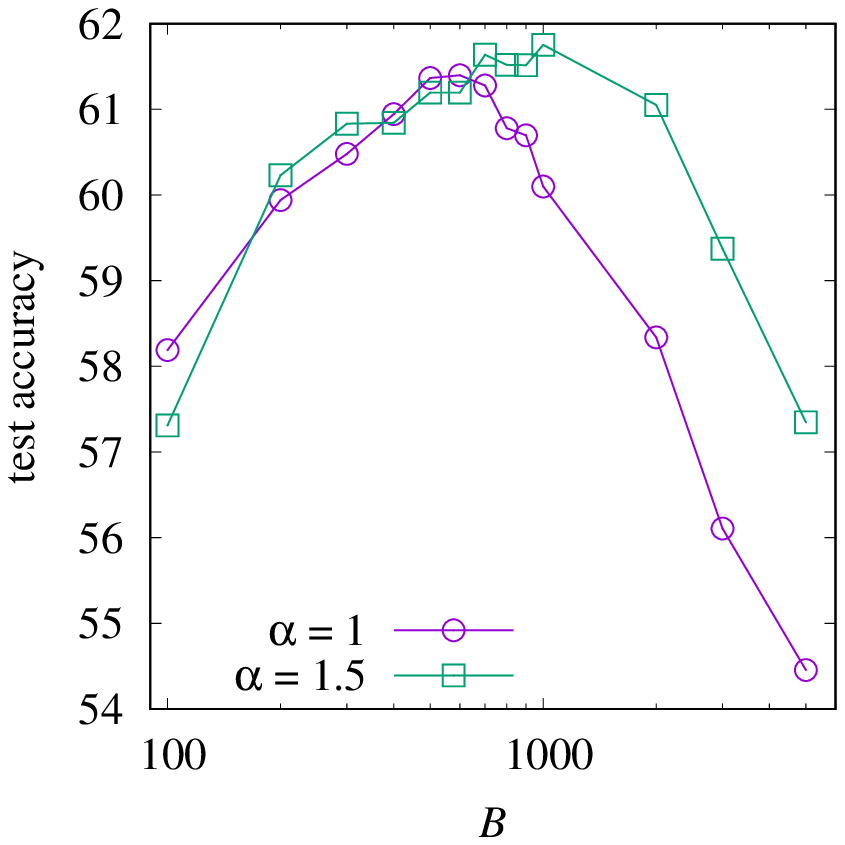}&
\includegraphics[width=0.5\linewidth]{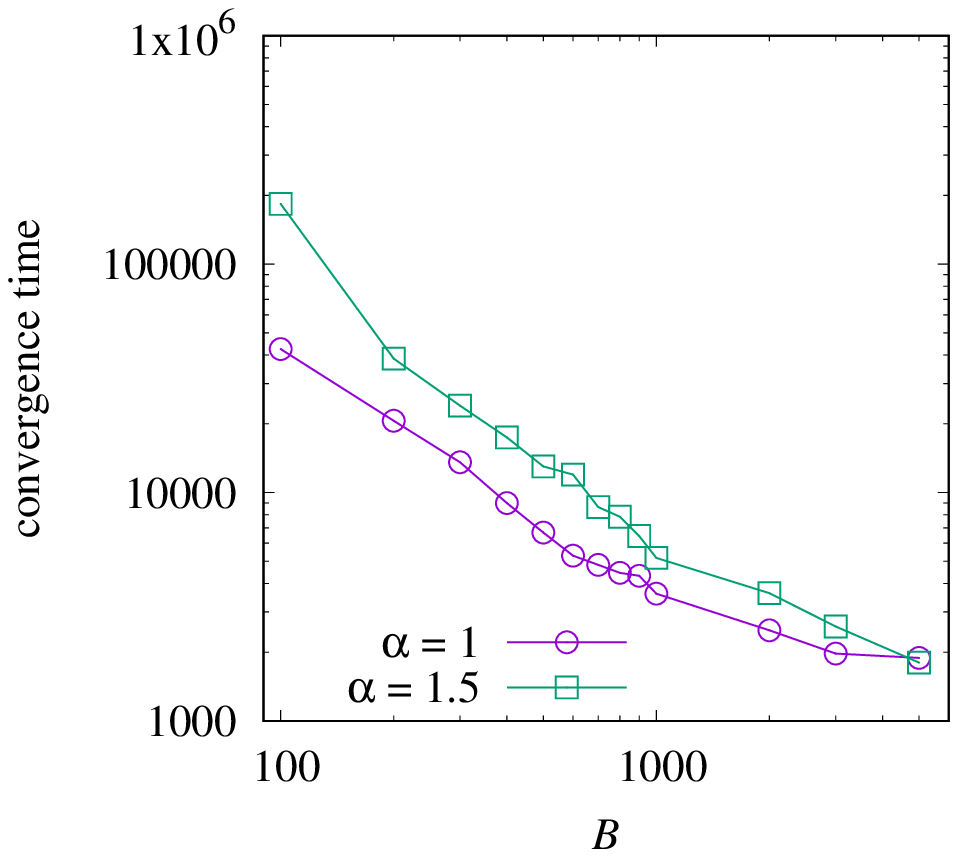}
\end{tabular}
\caption{Minibatch-size dependence of the test accuracy and the convergence time for each fixed value of $\alpha$ in F1 and C2.}
\label{fig:F1C2}
\end{figure}

Although we have focused on C1, other configurations F1 and C2 also show similar results.
For F1 and C2, we compare the result for $\alpha=1$ with that for $\alpha=1.5$.
In Fig.~\ref{fig:F1C2}, the minibatch-size dependences of the test accuracy and the convergence time are shown for F1 and C2.
In Table~\ref{t:batch_dep}, we also show the test accuracy and the convergence time at $B=B_\mathrm{opt}$ for each $\alpha$ in F1 and C2.
These results are qualitatively same as those in C1 (Fig.~\ref{fig:batch_dep} and Table~\ref{t:batch_dep}).

\subsection{Comparison between the noise enhancement and reducing the minibatch size}

It is pointed out that reducing the minibatch size $B$ with $\alpha=1$ has a similar effect as the NE with a fixed $B$; it results in better generalization but a longer convergence time\footnote{As was already mentioned, under the Gaussian noise approximation, increasing $\alpha$ is indeed equivalent to reducing $B$ to $B_\mathrm{eff}$ given by Eq.~(\ref{eq:B_eff}).}.
We shall compare the large-batch training with the NE to the small-batch training without the NE.
First we calculate the test accuracy and the convergence time for varying $B$ and a fixed $\alpha=1$ (no NE).
We then calculate the test accuracy for varying $\alpha> 1$ and a fixed $B=5000$, which corresponds to large minibatch training.
In other words, we compare the effect of the NE with that of reducing $B$.

\begin{figure}[t]
\centering
\begin{tabular}{ccc}
\hspace{-2cm}
\includegraphics[width=0.5\linewidth]{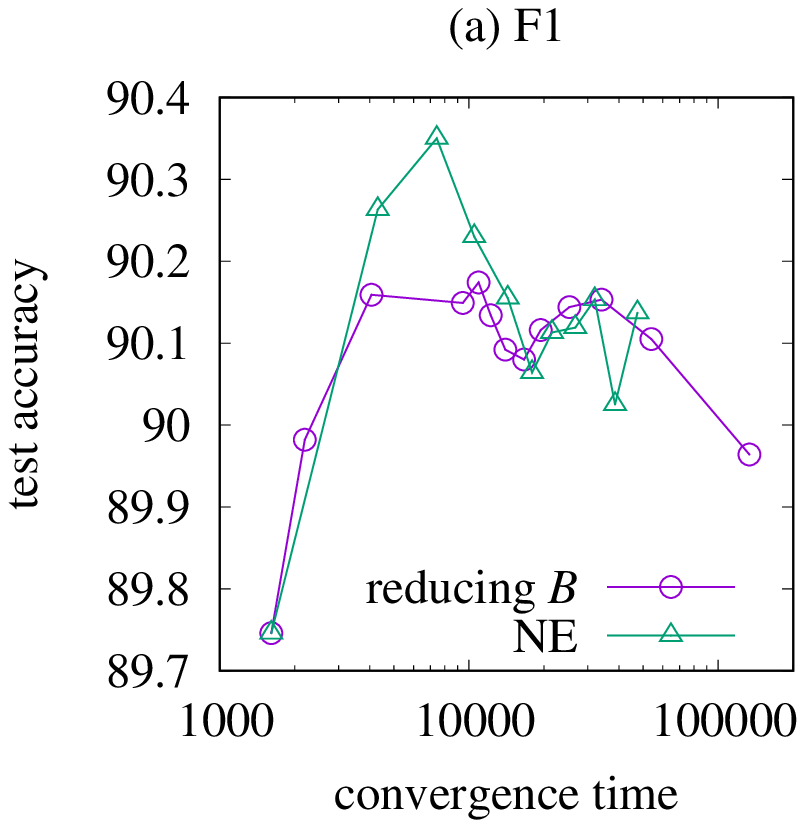}&
\hspace{-3cm}
\includegraphics[width=0.5\linewidth]{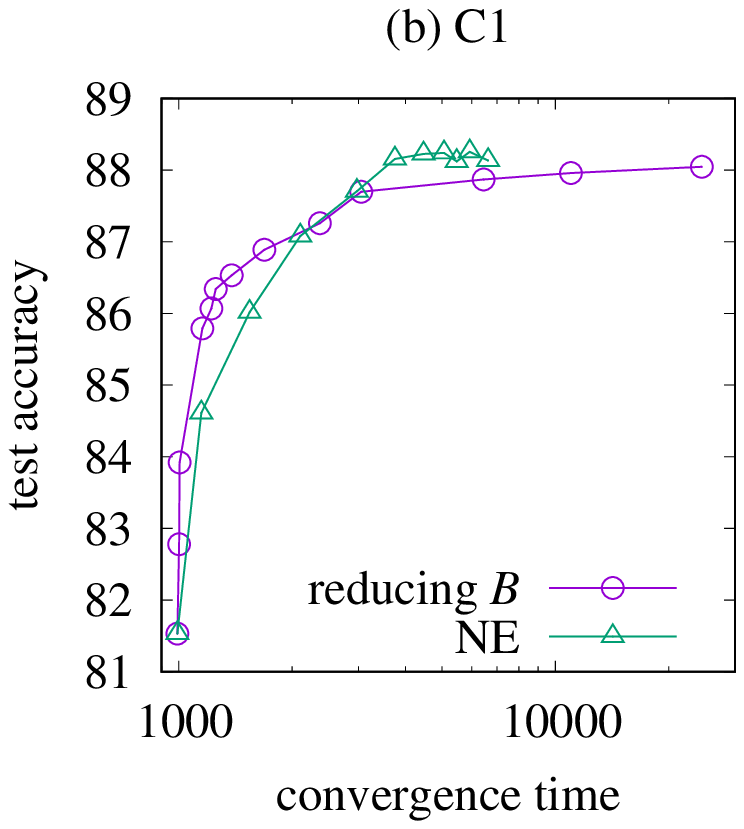}&
\hspace{-3cm}
\includegraphics[width=0.5\linewidth]{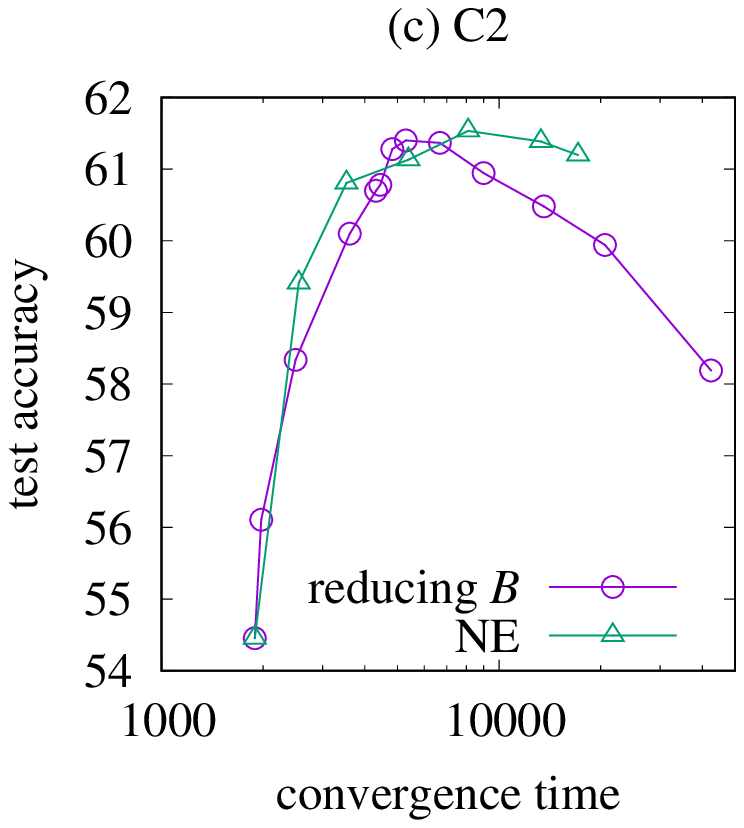}
\end{tabular}
\caption{Comparison between the effects of reducing the minibatch size $B$ with $\alpha=1$ and of increasing $\alpha$ with $B=5000$.
The longitudinal axis and the horizontal axis represent the test accuracy and the convergence time, respectively.
Circle data points (reducing $B$ with $\alpha=1$) correspond to $B=5000, 3000, 2000, 1000, 900, 800, 700, 600, 500, 400, 300, 200, 100$ from left to right.
Triangle data points (increasing $\alpha$ with $B=5000$) correspond to $\alpha=1,2,\dots, 11$ for F1 and C1, and $\alpha=1,2,\dots,7$ for C2, from left to right.}
\label{fig:data}
\end{figure}

The comparison between reducing $B$ with $\alpha=1$ and increasing $\alpha$ with $B=5000$ is given in Fig.~\ref{fig:data}.
We see that both give similar curves; increasing the convergence time with a peaked test accuracy.
However, in every case of F1, C1, and C2, the NE (increasing $\alpha$) results in better accuracy compared with reducing $B$ if $\alpha$ is properly chosen.

In Table~\ref{t:best}, we compare the best test accuracies between varying $B$ with $\alpha=1$ (without the NE) and increasing $\alpha$ with $B=5000$ (with the NE).
In all cases, the large-batch training with the NE outperforms the small-batch training without the NE.

\begin{table}[t]
\caption{Comparison of best test accuracies for varying $B$ with $\alpha=1$ (without the noise enhancement) and for varying $\alpha$ with $B=5000$ (with the noise enhancement).
The range of varying $B$ and $\alpha$ is the same as in Fig.~\ref{fig:data}.}
\vspace{11pt}
\centering
\begin{tabular}{lllll}
\hline
Name& & $B$ & $\alpha$ & Best test accuracy (\%) \\
\hline\hline
F1
&without NE & 900 & 1 & $90.17\pm 0.14$ \\
&with NE & 5000 & 3  & $\bm{90.35\pm 0.05}$\\
\hline
C1
&without NE & 100 & 1 & $88.05\pm 0.18$ \\
&with NE & 5000 & 10 & $\bm{88.26\pm 0.23}$ \\
\hline
C2
&without NE & 600 & 1 & $61.40\pm 0.54$ \\
&with NE & 5000 & 5 & $\bm{61.53\pm 0.35}$ \\
\hline
\end{tabular}
\label{t:best}
\end{table}

\section{Discussion}
\label{sec:discussion}

We have shown that the method of the NE for gradient-based optimization algorithms improves generalization.
In particular, large-batch training with the NE even outperforms small-batch training without the NE, which clearly shows that the NE is not equivalent to reducing the minibatch size $B$.

In this section, we shall discuss two fundamental questions raised here:
\begin{description}
\item[(i)] \textit{Why does a stronger SGD noise result in a better generalization?}
\item[(ii)] \textit{How is the inequivalence between the NE and reducing $B$ theoretically understood?}
\end{description}

We first consider (i).
When the SGD noise strength is inhomogeneous in the parameter space, network parameters will be likely to evolve to a minimum of the loss landscape with a weaker SGD noise\footnote{In physics, similar phenomena are known; Brownian particles in a medium with inhomogeneous temperature tend to gather in a colder region (Soret effect)~\citep{Duhr2006,Sancho2015}.}
That is, if the SGD noise is strong enough near a minimum, the network parameters will easily escape from it with the help of the SGD noise.
As a result, only minima around which the SGD noise is weak enough survive.
Since the covariance of the SGD noise is given by Eq.~(\ref{eq:noise_var}), or Eq.~(\ref{eq:NE_var}) for the NE, the strong SGD noise is considered to have an implicit regularization effect toward minima with a small variance of $\{\nabla_w\ell_\mu\}$.
Some previous studies have introduced various measures which express an implicit regularization effect of SGD~\citep{Keskar2017, Yin2018, Wu2018}.
Among them, the ``gradient diversity'' introduced by \citet{Yin2018} is closely related to the above argument.

A small variance of the sample-dependent gradients $\{\nabla_w\ell_\mu\}$ around a minimum of the loss function implies that the loss landscape $L_\mathcal{B}(w)$ for a minibatch $\mathcal{B}$ does not largely depend on $\mathcal{B}$.
Such a minimum would contain information on common features among training data samples, which would be relevant for a given classification, but not contain information on sample-specific features which lead to overfitting.
This is our intuitive picture that explains why the strong SGD noise results in good generalization performance.

The above consideration is solely based on Eq.~(\ref{eq:noise_var}), i.e., the covariance structure of the SGD noise, and the effect of non-Gaussian noise has been ignored.
However, when the SGD noise is strengthened by reducing $B$, the SGD noise deviates from Gaussian and the above argument should be somehow modified.
As we have already mentioned, the inequivalence between the NE and reducing $B$ results from the non-Gaussian nature of the SGD noise, which is therefore a key ingredient to answer the question (ii).
The method of the NE can increase the noise strength without changing $B$, and hence it is considered to suppress the non-Gaussianity compared with the case of just reducing $B$.
The experimental result presented in Sec.~\ref{sec:experiment} then indicates that the non-Gaussian nature of the SGD noise has a negative impact on generalization.
A possible interpretation is that sample-specific features show up and are overestimated, which results in overfitting, when the central limit theorem is strongly violated\footnote{At a certain stage of training, some training data samples have been confidently classified correctly but others have not. This fact suggests that the distributions of $\ell_\mu(w)$ and $\nabla_w\ell_\mu(w)$ have a long tail and that the variance of $\{\nabla_w\ell_\mu\}$ is not small enough to justify the central limit theorem unless $B$ is sufficiently large. Indeed, \citet{Panigrahi2019} have demonstrated that the SGD noise looks Gaussian only in an early stage of training for a not too large $B$.}.
However, the relation between the non-Gaussianity of the SGD noise and generalization remains unclear~\citep{Wu2019}, and it would be an important future problem to make this point clear.

In this way, we now have intuitive arguments which might be relevant to answer the questions (i) and (ii), but theoretical solid explanations are still lacking.
Our results will not only be useful in practice, but also give theoretical insights into those fundamental questions, which merit further study.

\bibliography{apsrevcontrol,deep_learning,physics}

\begin{thebibliography}{29}%
\makeatletter
\providecommand \@ifxundefined [1]{%
 \@ifx{#1\undefined}
}%
\providecommand \@ifnum [1]{%
 \ifnum #1\expandafter \@firstoftwo
 \else \expandafter \@secondoftwo
 \fi
}%
\providecommand \@ifx [1]{%
 \ifx #1\expandafter \@firstoftwo
 \else \expandafter \@secondoftwo
 \fi
}%
\providecommand \natexlab [1]{#1}%
\providecommand \enquote  [1]{``#1''}%
\providecommand \bibnamefont  [1]{#1}%
\providecommand \bibfnamefont [1]{#1}%
\providecommand \citenamefont [1]{#1}%
\providecommand \href@noop [0]{\@secondoftwo}%
\providecommand \href [0]{\begingroup \@sanitize@url \@href}%
\providecommand \@href[1]{\@@startlink{#1}\@@href}%
\providecommand \@@href[1]{\endgroup#1\@@endlink}%
\providecommand \@sanitize@url [0]{\catcode `\\12\catcode `\$12\catcode
  `\&12\catcode `\#12\catcode `\^12\catcode `\_12\catcode `\%12\relax}%
\providecommand \@@startlink[1]{}%
\providecommand \@@endlink[0]{}%
\providecommand \url  [0]{\begingroup\@sanitize@url \@url }%
\providecommand \@url [1]{\endgroup\@href {#1}{\urlprefix }}%
\providecommand \urlprefix  [0]{URL }%
\providecommand \Eprint [0]{\href }%
\providecommand \doibase [0]{https://doi.org/}%
\providecommand \selectlanguage [0]{\@gobble}%
\providecommand \bibinfo  [0]{\@secondoftwo}%
\providecommand \bibfield  [0]{\@secondoftwo}%
\providecommand \translation [1]{[#1]}%
\providecommand \BibitemOpen [0]{}%
\providecommand \bibitemStop [0]{}%
\providecommand \bibitemNoStop [0]{.\EOS\space}%
\providecommand \EOS [0]{\spacefactor3000\relax}%
\providecommand \BibitemShut  [1]{\csname bibitem#1\endcsname}%
\let\auto@bib@innerbib\@empty
\bibitem [{\citenamefont {Zhang}\ \emph {et~al.}(2017)\citenamefont {Zhang},
  \citenamefont {Bengio}, \citenamefont {Hardt}, \citenamefont {Recht},\ and\
  \citenamefont {Vinyals}}]{Zhang2017}%
  \BibitemOpen
  \bibfield  {author} {\bibinfo {author} {\bibfnamefont {Chiyuan}\ \bibnamefont
  {Zhang}}, \bibinfo {author} {\bibfnamefont {Samy}\ \bibnamefont {Bengio}},
  \bibinfo {author} {\bibfnamefont {Moritz}\ \bibnamefont {Hardt}}, \bibinfo
  {author} {\bibfnamefont {Benjamin}\ \bibnamefont {Recht}},\ and\ \bibinfo
  {author} {\bibfnamefont {Oriol}\ \bibnamefont {Vinyals}},\ }\bibfield
  {title} {\bibinfo {title} {{Understanding Deep Learning Requires Rethinking
  of Generalization}},\ }in\ \href
  {https://openreview.net/forum?id=Sy8gdB9xx&;amp;noteId=Sy8gdB9xx} {\emph
  {\bibinfo {booktitle} {International Conference on Learning
  Representations}}}\ (\bibinfo {year} {2017})\BibitemShut {NoStop}%
\bibitem [{\citenamefont {Dziugaite}\ and\ \citenamefont
  {Roy}(2017)}]{Dziugaite2017}%
  \BibitemOpen
  \bibfield  {author} {\bibinfo {author} {\bibfnamefont {Gintare~Karolina}\
  \bibnamefont {Dziugaite}}\ and\ \bibinfo {author} {\bibfnamefont {Daniel~M}\
  \bibnamefont {Roy}},\ }\bibfield  {title} {\bibinfo {title} {{Computing
  nonvacuous generalization bounds for deep (stochastic) neural networks with
  many more parameters than training data}},\ }in\ \href@noop {} {\emph
  {\bibinfo {booktitle} {Uncertainty in Artificial Intelligence}}}\ (\bibinfo
  {year} {2017})\ \Eprint {https://arxiv.org/abs/1703.11008} {arXiv:1703.11008}\BibitemShut {NoStop}%
\bibitem [{\citenamefont {Nagarajan}\ and\ \citenamefont
  {Kolter}(2017)}]{Nagarajan2017}%
  \BibitemOpen
  \bibfield  {author} {\bibinfo {author} {\bibfnamefont {Vaishnavh}\
  \bibnamefont {Nagarajan}}\ and\ \bibinfo {author} {\bibfnamefont {J.~Zico}\
  \bibnamefont {Kolter}},\ }\bibfield  {title} {\bibinfo {title}
  {{Generalization in Deep Networks: The Role of Distance from
  Initialization}},\ }in\ \href {http://arxiv.org/abs/1901.01672} {\emph
  {\bibinfo {booktitle} {Advances in Neural Information Processing Systems}}}\
  (\bibinfo {year} {2017})\ \Eprint {https://arxiv.org/abs/1901.01672}
  {arXiv:1901.01672}\BibitemShut {NoStop}%
\bibitem [{\citenamefont {Neyshabur}\ \emph {et~al.}(2017)\citenamefont
  {Neyshabur}, \citenamefont {Bhojanapalli}, \citenamefont {Mcallester},\ and\
  \citenamefont {Srebro}}]{Neyshabur2017}%
  \BibitemOpen
  \bibfield  {author} {\bibinfo {author} {\bibfnamefont {Behnam}\ \bibnamefont
  {Neyshabur}}, \bibinfo {author} {\bibfnamefont {Srinadh}\ \bibnamefont
  {Bhojanapalli}}, \bibinfo {author} {\bibfnamefont {David}\ \bibnamefont
  {Mcallester}},\ and\ \bibinfo {author} {\bibfnamefont {Nathan}\ \bibnamefont
  {Srebro}},\ }\bibfield  {title} {\bibinfo {title} {{Exploring Generalization
  in Deep Learning}},\ }in\ \href
  {http://papers.nips.cc/paper/7176-exploring-generalization-in-deep-learning}
  {\emph {\bibinfo {booktitle} {Advances in Neural Information Processing
  Systems}}}\ (\bibinfo {year} {2017})\BibitemShut {NoStop}%
\bibitem [{\citenamefont {Neyshabur}\ \emph {et~al.}(2019)\citenamefont
  {Neyshabur}, \citenamefont {Li}, \citenamefont {Bhojanapalli}, \citenamefont
  {LeCun},\ and\ \citenamefont {Srebro}}]{Neyshabur2019}%
  \BibitemOpen
  \bibfield  {author} {\bibinfo {author} {\bibfnamefont {Behnam}\ \bibnamefont
  {Neyshabur}}, \bibinfo {author} {\bibfnamefont {Zhiyuan}\ \bibnamefont {Li}},
  \bibinfo {author} {\bibfnamefont {Srinadh}\ \bibnamefont {Bhojanapalli}},
  \bibinfo {author} {\bibfnamefont {Yann}\ \bibnamefont {LeCun}},\ and\
  \bibinfo {author} {\bibfnamefont {Nathan}\ \bibnamefont {Srebro}},\
  }\bibfield  {title} {\bibinfo {title} {{The role of over-parametrization in
  generalization of neural networks}},\ }in\ \href
  {https://openreview.net/forum?id=BygfghAcYX} {\emph {\bibinfo {booktitle}
  {International Conference on Learning Representations}}}\ (\bibinfo {year}
  {2019})\ \Eprint {https://arxiv.org/abs/1805.12076} {arXiv:1805.12076}\BibitemShut {NoStop}%
\bibitem [{\citenamefont {Arora}\ \emph {et~al.}(2018)\citenamefont {Arora},
  \citenamefont {Ge}, \citenamefont {Neyshabur},\ and\ \citenamefont
  {Zhang}}]{Arora2018}%
  \BibitemOpen
  \bibfield  {author} {\bibinfo {author} {\bibfnamefont {Sanjeev}\ \bibnamefont
  {Arora}}, \bibinfo {author} {\bibfnamefont {Rong}\ \bibnamefont {Ge}},
  \bibinfo {author} {\bibfnamefont {Behnam}\ \bibnamefont {Neyshabur}},\ and\
  \bibinfo {author} {\bibfnamefont {Yi}~\bibnamefont {Zhang}},\ }\bibfield
  {title} {\bibinfo {title} {{Stronger generalization bounds for deep nets via
  a compression approach}},\ }in\ \href
  {http://proceedings.mlr.press/v80/arora18b.html} {\emph {\bibinfo {booktitle}
  {International Conference on Machine Learning}}}\ (\bibinfo {year} {2018})\
  \Eprint {https://arxiv.org/abs/1802.05296} {arXiv:1802.05296}\BibitemShut
  {NoStop}%
\bibitem [{\citenamefont {P{\'{e}}rez}\ \emph {et~al.}(2019)\citenamefont
  {P{\'{e}}rez}, \citenamefont {Louis},\ and\ \citenamefont
  {Camargo}}]{Perez2019}%
  \BibitemOpen
  \bibfield  {author} {\bibinfo {author} {\bibfnamefont {Guillermo~Valle}\
  \bibnamefont {P{\'{e}}rez}}, \bibinfo {author} {\bibfnamefont {Ard~A}\
  \bibnamefont {Louis}},\ and\ \bibinfo {author} {\bibfnamefont {Chico~Q}\
  \bibnamefont {Camargo}},\ }\bibfield  {title} {\bibinfo {title} {{Deep
  learning generalizes because the parameter-function map is biased towards
  simple functions}},\ }in\ \href {https://openreview.net/forum?id=rye4g3AqFm}
  {\emph {\bibinfo {booktitle} {International Conference on Learning
  Representations}}}\ (\bibinfo {year} {2019})\ \Eprint
  {https://arxiv.org/abs/1805.08522} {arXiv:1805.08522}\BibitemShut {NoStop}%
\bibitem [{\citenamefont {Jacot}\ \emph {et~al.}(2018)\citenamefont {Jacot},
  \citenamefont {Gabriel},\ and\ \citenamefont {Hongler}}]{Jacot2018}%
  \BibitemOpen
  \bibfield  {author} {\bibinfo {author} {\bibfnamefont {Arthur}\ \bibnamefont
  {Jacot}}, \bibinfo {author} {\bibfnamefont {Franck}\ \bibnamefont
  {Gabriel}},\ and\ \bibinfo {author} {\bibfnamefont {Cl{\'{e}}ment}\
  \bibnamefont {Hongler}},\ }\bibfield  {title} {\bibinfo {title} {{Neural
  tangent kernel: Convergence and generalization in neural networks}},\ }in\
  \href
  {http://papers.nips.cc/paper/8076-neural-tangent-kernel-convergence-and-generalization-in-neural-networks}
  {\emph {\bibinfo {booktitle} {Advances in Neural Information Processing
  Systems}}}\ (\bibinfo {year} {2018})\ \Eprint
  {https://arxiv.org/abs/1806.07572} {arXiv:1806.07572}\BibitemShut {NoStop}%
\bibitem [{\citenamefont {Arora}\ \emph {et~al.}(2019)\citenamefont {Arora},
  \citenamefont {Du}, \citenamefont {Hu}, \citenamefont {Li}, \citenamefont
  {Salakhutdinov},\ and\ \citenamefont {Wang}}]{Arora2019}%
  \BibitemOpen
  \bibfield  {author} {\bibinfo {author} {\bibfnamefont {Sanjeev}\ \bibnamefont
  {Arora}}, \bibinfo {author} {\bibfnamefont {Simon~S.}\ \bibnamefont {Du}},
  \bibinfo {author} {\bibfnamefont {Wei}\ \bibnamefont {Hu}}, \bibinfo {author}
  {\bibfnamefont {Zhiyuan}\ \bibnamefont {Li}}, \bibinfo {author}
  {\bibfnamefont {Ruslan}\ \bibnamefont {Salakhutdinov}},\ and\ \bibinfo
  {author} {\bibfnamefont {Ruosong}\ \bibnamefont {Wang}},\ }\bibfield  {title}
  {\bibinfo {title} {{On Exact Computation with an Infinitely Wide Neural
  Net}},\ }in\ \href {http://arxiv.org/abs/1904.11955} {\emph {\bibinfo
  {booktitle} {Neural Information Processing Systems}}}\ (\bibinfo {year}
  {2019})\ \Eprint {https://arxiv.org/abs/1904.11955} {arXiv:1904.11955}\BibitemShut {NoStop}%
\bibitem [{\citenamefont {D'Ascoli}\ \emph {et~al.}()\citenamefont {D'Ascoli},
  \citenamefont {Refinetti}, \citenamefont {Biroli},\ and\ \citenamefont
  {Krzakala}}]{dAscoli2020}%
  \BibitemOpen
  \bibfield  {author} {\bibinfo {author} {\bibfnamefont {St{\'{e}}phane}\
  \bibnamefont {D'Ascoli}}, \bibinfo {author} {\bibfnamefont {Maria}\
  \bibnamefont {Refinetti}}, \bibinfo {author} {\bibfnamefont {Giulio}\
  \bibnamefont {Biroli}},\ and\ \bibinfo {author} {\bibfnamefont {Florent}\
  \bibnamefont {Krzakala}},\ }\href@noop {} {\bibinfo {title} {{Double Trouble
  in Double Descent : Bias and Variance(s) in the Lazy Regime}}},\ \Eprint
  {https://arxiv.org/abs/2003.01054} {arXiv:2003.01054}\BibitemShut {NoStop}%
\bibitem [{\citenamefont {Zhu}\ \emph {et~al.}(2019)\citenamefont {Zhu},
  \citenamefont {Wu}, \citenamefont {Yu}, \citenamefont {Wu},\ and\
  \citenamefont {Ma}}]{Zhu2019}%
  \BibitemOpen
  \bibfield  {author} {\bibinfo {author} {\bibfnamefont {Zhanxing}\
  \bibnamefont {Zhu}}, \bibinfo {author} {\bibfnamefont {Jingfeng}\
  \bibnamefont {Wu}}, \bibinfo {author} {\bibfnamefont {Bing}\ \bibnamefont
  {Yu}}, \bibinfo {author} {\bibfnamefont {Lei}\ \bibnamefont {Wu}},\ and\
  \bibinfo {author} {\bibfnamefont {Jinwen}\ \bibnamefont {Ma}},\ }\bibfield
  {title} {\bibinfo {title} {{The anisotropic noise in stochastic gradient
  descent: Its behavior of escaping from sharp minima and regularization
  effects}},\ }in\ \href {http://proceedings.mlr.press/v97/zhu19e.html} {\emph
  {\bibinfo {booktitle} {International Conference on Machine Learning}}}\
  (\bibinfo {year} {2019})\ \Eprint {https://arxiv.org/abs/1803.00195}
  {arXiv:1803.00195}\BibitemShut {NoStop}%
\bibitem [{\citenamefont {Wu}\ \emph {et~al.}()\citenamefont {Wu},
  \citenamefont {Hu}, \citenamefont {Xiong}, \citenamefont {Huan},
  \citenamefont {Braverman},\ and\ \citenamefont {Zhu}}]{Wu2019}%
  \BibitemOpen
  \bibfield  {author} {\bibinfo {author} {\bibfnamefont {Jingfeng}\
  \bibnamefont {Wu}}, \bibinfo {author} {\bibfnamefont {Wenqing}\ \bibnamefont
  {Hu}}, \bibinfo {author} {\bibfnamefont {Haoyi}\ \bibnamefont {Xiong}},
  \bibinfo {author} {\bibfnamefont {Jun}\ \bibnamefont {Huan}}, \bibinfo
  {author} {\bibfnamefont {Vladimir}\ \bibnamefont {Braverman}},\ and\ \bibinfo
  {author} {\bibfnamefont {Zhanxing}\ \bibnamefont {Zhu}},\ }\href@noop {}
  {\bibinfo {title} {{On the Noisy Gradient Descent that Generalizes as
  SGD}}},\ \Eprint {https://arxiv.org/abs/1906.07405} {arXiv:1906.07405}\BibitemShut {NoStop}%
\bibitem [{\citenamefont {Smith}\ \emph {et~al.}(2020)\citenamefont {Smith},
  \citenamefont {Elsen},\ and\ \citenamefont {De}}]{Smith2020}%
  \BibitemOpen
  \bibfield  {author} {\bibinfo {author} {\bibfnamefont {Samuel~L.}\
  \bibnamefont {Smith}}, \bibinfo {author} {\bibfnamefont {Erich}\ \bibnamefont
  {Elsen}},\ and\ \bibinfo {author} {\bibfnamefont {Soham}\ \bibnamefont
  {De}},\ }\bibfield  {title} {\bibinfo {title} {{On the Generalization Benefit
  of Noise in Stochastic Gradient Descent}},\ }\href
  {http://arxiv.org/abs/2006.15081} {\  (\bibinfo {year} {2020})},\ \Eprint
  {https://arxiv.org/abs/2006.15081} {arXiv:2006.15081}\BibitemShut {NoStop}%
\bibitem [{\citenamefont {Li}\ \emph {et~al.}(2017)\citenamefont {Li},
  \citenamefont {Tai},\ and\ \citenamefont {Weinan}}]{Li2017}%
  \BibitemOpen
  \bibfield  {author} {\bibinfo {author} {\bibfnamefont {Qianxiao}\
  \bibnamefont {Li}}, \bibinfo {author} {\bibfnamefont {Cheng}\ \bibnamefont
  {Tai}},\ and\ \bibinfo {author} {\bibfnamefont {E.}~\bibnamefont {Weinan}},\
  }\bibfield  {title} {\bibinfo {title} {{Stochastic modified equations and
  adaptive stochastic gradient algorithms}},\ }in\ \href
  {http://proceedings.mlr.press/v70/li17f.html} {\emph {\bibinfo {booktitle}
  {International Conference on Machine Learning}}}\ (\bibinfo {year} {2017})\
  \Eprint {https://arxiv.org/abs/1511.06251} {arXiv:1511.06251}\BibitemShut
  {NoStop}%
\bibitem [{\citenamefont {Jastrz{\c{e}}bski}\ \emph {et~al.}()\citenamefont
  {Jastrz{\c{e}}bski}, \citenamefont {Kenton}, \citenamefont {Arpit},
  \citenamefont {Ballas}, \citenamefont {Fischer}, \citenamefont {Bengio},\
  and\ \citenamefont {Storkey}}]{Jastrzebski2017}%
  \BibitemOpen
  \bibfield  {author} {\bibinfo {author} {\bibfnamefont {Stanis{\l}aw}\
  \bibnamefont {Jastrz{\c{e}}bski}}, \bibinfo {author} {\bibfnamefont
  {Zachary}\ \bibnamefont {Kenton}}, \bibinfo {author} {\bibfnamefont
  {Devansh}\ \bibnamefont {Arpit}}, \bibinfo {author} {\bibfnamefont {Nicolas}\
  \bibnamefont {Ballas}}, \bibinfo {author} {\bibfnamefont {Asja}\ \bibnamefont
  {Fischer}}, \bibinfo {author} {\bibfnamefont {Yoshua}\ \bibnamefont
  {Bengio}},\ and\ \bibinfo {author} {\bibfnamefont {Amos}\ \bibnamefont
  {Storkey}},\ }\href@noop {} {\bibinfo {title} {{Three Factors Influencing
  Minima in SGD}}},\ \Eprint {https://arxiv.org/abs/1711.04623}
  {arXiv:1711.04623}\BibitemShut {NoStop}%
\bibitem [{\citenamefont {Goyal}\ \emph {et~al.}()\citenamefont {Goyal},
  \citenamefont {Doll{\'{a}}r}, \citenamefont {Girshick}, \citenamefont
  {Noordhuis}, \citenamefont {Wesolowski}, \citenamefont {Kyrola},
  \citenamefont {Tulloch}, \citenamefont {Jia},\ and\ \citenamefont
  {He}}]{Goyal2017}%
  \BibitemOpen
  \bibfield  {author} {\bibinfo {author} {\bibfnamefont {Priya}\ \bibnamefont
  {Goyal}}, \bibinfo {author} {\bibfnamefont {Piotr}\ \bibnamefont
  {Doll{\'{a}}r}}, \bibinfo {author} {\bibfnamefont {Ross}\ \bibnamefont
  {Girshick}}, \bibinfo {author} {\bibfnamefont {Pieter}\ \bibnamefont
  {Noordhuis}}, \bibinfo {author} {\bibfnamefont {Lukasz}\ \bibnamefont
  {Wesolowski}}, \bibinfo {author} {\bibfnamefont {Aapo}\ \bibnamefont
  {Kyrola}}, \bibinfo {author} {\bibfnamefont {Andrew}\ \bibnamefont
  {Tulloch}}, \bibinfo {author} {\bibfnamefont {Yangqing}\ \bibnamefont
  {Jia}},\ and\ \bibinfo {author} {\bibfnamefont {Kaiming}\ \bibnamefont
  {He}},\ }\href@noop {} {\bibinfo {title} {{Accurate, Large Minibatch SGD:
  Training ImageNet in 1 Hour}}},\ \Eprint {https://arxiv.org/abs/1706.02677}
  {arXiv:1706.02677}\BibitemShut {NoStop}%
\bibitem [{\citenamefont {Smith}\ and\ \citenamefont
  {Le}(2018)}]{Smith-Le2018}%
  \BibitemOpen
  \bibfield  {author} {\bibinfo {author} {\bibfnamefont {Samuel~L.}\
  \bibnamefont {Smith}}\ and\ \bibinfo {author} {\bibfnamefont {Quoc~V.}\
  \bibnamefont {Le}},\ }\bibfield  {title} {\bibinfo {title} {{A Bayesian
  perspective on generalization and stochastic gradient descent}},\ }in\ \href
  {https://openreview.net/forum?id=BJij4yg0Z} {\emph {\bibinfo {booktitle}
  {International Conference on Learning Representations}}}\ (\bibinfo {year}
  {2018})\ \Eprint {https://arxiv.org/abs/1710.06451} {arXiv:1710.06451}\BibitemShut {NoStop}%
\bibitem [{\citenamefont {Hoffer}\ \emph {et~al.}(2017)\citenamefont {Hoffer},
  \citenamefont {Hubara},\ and\ \citenamefont {Soudry}}]{Hoffer2017}%
  \BibitemOpen
  \bibfield  {author} {\bibinfo {author} {\bibfnamefont {Elad}\ \bibnamefont
  {Hoffer}}, \bibinfo {author} {\bibfnamefont {Itay}\ \bibnamefont {Hubara}},\
  and\ \bibinfo {author} {\bibfnamefont {Daniel}\ \bibnamefont {Soudry}},\
  }\bibfield  {title} {\bibinfo {title} {{Train longer, generalize better:
  closing the generalization gap in large batch training of neural networks}},\
  }in\ \href {https://doi.org/10.1016/j.jcjd.2014.02.001} {\emph {\bibinfo
  {booktitle} {Advances in Neural Information Processing Systems}}}\ (\bibinfo
  {year} {2017})\ \Eprint {https://arxiv.org/abs/1705.08741} {arXiv:1705.08741}\BibitemShut {NoStop}%
\bibitem [{\citenamefont {Hoffer}\ \emph {et~al.}()\citenamefont {Hoffer},
  \citenamefont {Ben-Nun}, \citenamefont {Hubara}, \citenamefont {Giladi},
  \citenamefont {Hoefler},\ and\ \citenamefont {Soudry}}]{Hoffer2019}%
  \BibitemOpen
  \bibfield  {author} {\bibinfo {author} {\bibfnamefont {Elad}\ \bibnamefont
  {Hoffer}}, \bibinfo {author} {\bibfnamefont {Tal}\ \bibnamefont {Ben-Nun}},
  \bibinfo {author} {\bibfnamefont {Itay}\ \bibnamefont {Hubara}}, \bibinfo
  {author} {\bibfnamefont {Niv}\ \bibnamefont {Giladi}}, \bibinfo {author}
  {\bibfnamefont {Torsten}\ \bibnamefont {Hoefler}},\ and\ \bibinfo {author}
  {\bibfnamefont {Daniel}\ \bibnamefont {Soudry}},\ }\href@noop {} {\bibinfo
  {title} {{Augment your batch: better training with larger batches}}},\
  \Eprint {https://arxiv.org/abs/1901.09335} {arXiv:1901.09335}\BibitemShut
  {NoStop}%
\bibitem [{\citenamefont {Keskar}\ \emph {et~al.}(2017)\citenamefont {Keskar},
  \citenamefont {Nocedal}, \citenamefont {Tang}, \citenamefont {Mudigere},\
  and\ \citenamefont {Smelyanskiy}}]{Keskar2017}%
  \BibitemOpen
  \bibfield  {author} {\bibinfo {author} {\bibfnamefont {Nitish~Shirish}\
  \bibnamefont {Keskar}}, \bibinfo {author} {\bibfnamefont {Jorge}\
  \bibnamefont {Nocedal}}, \bibinfo {author} {\bibfnamefont {Ping Tak~Peter}\
  \bibnamefont {Tang}}, \bibinfo {author} {\bibfnamefont {Dheevatsa}\
  \bibnamefont {Mudigere}},\ and\ \bibinfo {author} {\bibfnamefont {Mikhail}\
  \bibnamefont {Smelyanskiy}},\ }\bibfield  {title} {\bibinfo {title} {{On
  large-batch training for deep learning: Generalization gap and sharp
  minima}},\ }in\ \href {https://openreview.net/forum?id=H1oyRlYgg} {\emph
  {\bibinfo {booktitle} {International Conference on Learning
  Representations}}}\ (\bibinfo {year} {2017})\ \Eprint
  {https://arxiv.org/abs/1609.04836} {arXiv:1609.04836}\BibitemShut {NoStop}%
\bibitem [{\citenamefont {Wu}\ \emph {et~al.}(2018)\citenamefont {Wu},
  \citenamefont {Ma},\ and\ \citenamefont {Weinan}}]{Wu2018}%
  \BibitemOpen
  \bibfield  {author} {\bibinfo {author} {\bibfnamefont {Lei}\ \bibnamefont
  {Wu}}, \bibinfo {author} {\bibfnamefont {Chao}\ \bibnamefont {Ma}},\ and\
  \bibinfo {author} {\bibfnamefont {E.}~\bibnamefont {Weinan}},\ }\bibfield
  {title} {\bibinfo {title} {{How SGD selects the global minima in
  over-parameterized learning: A dynamical stability perspective}},\ }in\ \href
  {http://papers.nips.cc/paper/8049-how-sgd-selects-the-global-minima-in-over-parameterized-learning-a-dynamical-stability-perspective}
  {\emph {\bibinfo {booktitle} {Advances in Neural Information Processing
  Systems}}}\ (\bibinfo {year} {2018})\BibitemShut {NoStop}%
\bibitem [{\citenamefont {Golmant}\ \emph {et~al.}()\citenamefont {Golmant},
  \citenamefont {Vemuri}, \citenamefont {Yao}, \citenamefont {Feinberg},
  \citenamefont {Gholami}, \citenamefont {Rothauge}, \citenamefont {Mahoney},\
  and\ \citenamefont {Gonzalez}}]{Golmant2018}%
  \BibitemOpen
  \bibfield  {author} {\bibinfo {author} {\bibfnamefont {Noah}\ \bibnamefont
  {Golmant}}, \bibinfo {author} {\bibfnamefont {Nikita}\ \bibnamefont
  {Vemuri}}, \bibinfo {author} {\bibfnamefont {Zhewei}\ \bibnamefont {Yao}},
  \bibinfo {author} {\bibfnamefont {Vladimir}\ \bibnamefont {Feinberg}},
  \bibinfo {author} {\bibfnamefont {Amir}\ \bibnamefont {Gholami}}, \bibinfo
  {author} {\bibfnamefont {Kai}\ \bibnamefont {Rothauge}}, \bibinfo {author}
  {\bibfnamefont {Michael~W.}\ \bibnamefont {Mahoney}},\ and\ \bibinfo {author}
  {\bibfnamefont {Joseph}\ \bibnamefont {Gonzalez}},\ }\href@noop {} {\bibinfo
  {title} {{On the Computational Inefficiency of Large Batch Sizes for
  Stochastic Gradient Descent}}},\ \Eprint {https://arxiv.org/abs/1811.12941}
  {arXiv:1811.12941}\BibitemShut {NoStop}%
\bibitem [{\citenamefont {Krizhevsky}()}]{Krizhevsky2014}%
  \BibitemOpen
  \bibfield  {author} {\bibinfo {author} {\bibfnamefont {Alex}\ \bibnamefont
  {Krizhevsky}},\ }\href@noop {} {\bibinfo {title} {{One weird trick for
  parallelizing convolutional neural networks}}},\ \Eprint
  {https://arxiv.org/abs/1404.5997} {arXiv:1404.5997}\BibitemShut {NoStop}%
\bibitem [{\citenamefont {Smith}\ \emph {et~al.}(2018)\citenamefont {Smith},
  \citenamefont {Kindermans}, \citenamefont {Ying},\ and\ \citenamefont
  {Le}}]{Smith2018}%
  \BibitemOpen
  \bibfield  {author} {\bibinfo {author} {\bibfnamefont {Samuel~L}\
  \bibnamefont {Smith}}, \bibinfo {author} {\bibfnamefont {Pieter~Jan}\
  \bibnamefont {Kindermans}}, \bibinfo {author} {\bibfnamefont {Chris}\
  \bibnamefont {Ying}},\ and\ \bibinfo {author} {\bibfnamefont {Quoc~V}\
  \bibnamefont {Le}},\ }\bibfield  {title} {\bibinfo {title} {{Don't decay the
  learning rate, increase the batch size}},\ }in\ \href
  {https://openreview.net/forum?id=B1Yy1BxCZ&noteId=B1Yy1BxCZ} {\emph {\bibinfo
  {booktitle} {International Conference on Learning Representations}}}\
  (\bibinfo {year} {2018})\ \Eprint {https://arxiv.org/abs/1711.00489}
  {arXiv:1711.00489}\BibitemShut {NoStop}%
\bibitem [{\citenamefont {Panigrahi}\ \emph {et~al.}()\citenamefont
  {Panigrahi}, \citenamefont {Somani}, \citenamefont {Goyal},\ and\
  \citenamefont {Netrapalli}}]{Panigrahi2019}%
  \BibitemOpen
  \bibfield  {author} {\bibinfo {author} {\bibfnamefont {Abhishek}\
  \bibnamefont {Panigrahi}}, \bibinfo {author} {\bibfnamefont {Raghav}\
  \bibnamefont {Somani}}, \bibinfo {author} {\bibfnamefont {Navin}\
  \bibnamefont {Goyal}},\ and\ \bibinfo {author} {\bibfnamefont {Praneeth}\
  \bibnamefont {Netrapalli}},\ }\href@noop {} {\bibinfo {title}
  {{Non-Gaussianity of Stochastic Gradient Noise}}},\ \Eprint
  {https://arxiv.org/abs/1910.09626} {arXiv:1910.09626}\BibitemShut {NoStop}%
\bibitem [{\citenamefont {Simonyan}\ and\ \citenamefont
  {Zisserman}()}]{Simonyan2014}%
  \BibitemOpen
  \bibfield  {author} {\bibinfo {author} {\bibfnamefont {Karen}\ \bibnamefont
  {Simonyan}}\ and\ \bibinfo {author} {\bibfnamefont {Andrew}\ \bibnamefont
  {Zisserman}},\ }\href@noop {} {\bibinfo {title} {{Very deep convolutional
  networks for large-scale image recognition}}},\ \Eprint
  {https://arxiv.org/abs/1409.1556v6} {arXiv:1409.1556v6}\BibitemShut
  {NoStop}%
\bibitem [{\citenamefont {Duhr}\ and\ \citenamefont {Braun}(2006)}]{Duhr2006}%
  \BibitemOpen
  \bibfield  {author} {\bibinfo {author} {\bibfnamefont {Stefan}\ \bibnamefont
  {Duhr}}\ and\ \bibinfo {author} {\bibfnamefont {Dieter}\ \bibnamefont
  {Braun}},\ }\bibfield  {title} {\bibinfo {title} {{Thermophoretic depletion
  follows boltzmann distribution}},\ }\href
  {https://doi.org/10.1103/PhysRevLett.96.168301} {\bibfield  {journal}
  {\bibinfo  {journal} {Physical Review Letters}\ }\textbf {\bibinfo {volume}
  {96}},\ \bibinfo {pages} {168301} (\bibinfo {year} {2006})}\BibitemShut
  {NoStop}%
\bibitem [{\citenamefont {Sancho}(2015)}]{Sancho2015}%
  \BibitemOpen
  \bibfield  {author} {\bibinfo {author} {\bibfnamefont {J.~M.}\ \bibnamefont
  {Sancho}},\ }\bibfield  {title} {\bibinfo {title} {{Brownian colloids in
  underdamped and overdamped regimes with nonhomogeneous temperature}},\ }\href
  {https://doi.org/10.1103/PhysRevE.92.062110} {\bibfield  {journal} {\bibinfo
  {journal} {Physical Review E}\ }\textbf {\bibinfo {volume} {92}},\ \bibinfo
  {pages} {062110} (\bibinfo {year} {2015})}\BibitemShut {NoStop}%
\bibitem [{\citenamefont {Yin}\ \emph {et~al.}(2018)\citenamefont {Yin},
  \citenamefont {Pananjady}, \citenamefont {Lam}, \citenamefont
  {Papailiopoulos}, \citenamefont {Ramchandran},\ and\ \citenamefont
  {Bartlett}}]{Yin2018}%
  \BibitemOpen
  \bibfield  {author} {\bibinfo {author} {\bibfnamefont {Dong}\ \bibnamefont
  {Yin}}, \bibinfo {author} {\bibfnamefont {Ashwin}\ \bibnamefont {Pananjady}},
  \bibinfo {author} {\bibfnamefont {Max}\ \bibnamefont {Lam}}, \bibinfo
  {author} {\bibfnamefont {Dimitris}\ \bibnamefont {Papailiopoulos}}, \bibinfo
  {author} {\bibfnamefont {Kannan}\ \bibnamefont {Ramchandran}},\ and\ \bibinfo
  {author} {\bibfnamefont {Peter~L.}\ \bibnamefont {Bartlett}},\ }\bibfield
  {title} {\bibinfo {title} {{Gradient diversity: A key ingredient for scalable
  distributed learning}},\ }in\ \href
  {http://proceedings.mlr.press/v84/yin18a.html} {\emph {\bibinfo {booktitle}
  {International Conference on Artificial Intelligence and Statistics}}}\
  (\bibinfo {year} {2018})\ \Eprint {https://arxiv.org/abs/1706.05699}
  {arXiv:1706.05699}\BibitemShut {NoStop}%
\end{thebibliography}%

\end{document}